\documentclass{opt2026} % Include author names
\jmlrproceedings{Preprint}{Preprint. Under review.}

\usepackage[utf8]{inputenc}
\usepackage{amsmath,amssymb,mathtools}
\usepackage{microtype}
\usepackage{sansmath}
\usepackage{xcolor}
\usepackage{booktabs}
\usepackage{xurl}
\usepackage{placeins}
\usepackage{hyperref}

\newcommand{\R}{\mathbb{R}}
\newcommand{\one}{\mathbf{1}}
\newcommand{\norm}[1]{\left\lVert #1\right\rVert}
\newcommand{\abs}[1]{\left\lvert #1\right\rvert}
\newcommand{\inner}[2]{\left\langle #1,#2\right\rangle}
\newcommand{\varnorm}[1]{\left\lVert #1\right\rVert_{\mathrm{var}}}
\DeclareMathOperator{\softmax}{softmax}
\DeclareMathOperator{\rank}{rank}

\title{Toward a First-Principles Update Geometry
for the Language-Model Head}

\optauthor{%
\Name{Aditya Somasundaram} \Email{aditya.somasundaram@mbzuai.ac.ae}\\
\Name{Charles Guille-Escuret} \Email{C.GuilleEscuret@mbzuai.ac.ae}\\
\Name{Alexander Moreno} \Email{alex.moreno@mbzuai.ac.ae} \\
\Name{Zhengzhong Liu} \Email{hector.liu@mbzuai.ac.ae} \\
\Name{Eric Xing} \Email{eric.xing@mbzuai.ac.ae} \\
}

\begin{document}

\maketitle

\begin{abstract}
Muon motivates designing optimizer geometry around the function of each parameter block and uses the spectral norm for hidden linear layers. 
For the language-model head, the spectral norm is not a faithful measure of functional change.
Softmax removes shared logit shifts, whereas the spectral norm can assign arbitrarily large size to updates that change no output probability.
We therefore treat the LM head and softmax as one module and derive an update geometry for their composition.
Hilbert's projective distance respects this invariance as it measures the largest change in pairwise log odds.
For an update $S$ with token rows $s_i^\top$, we show that the largest Hilbert distance over $\norm{h}_2\leq H$ is exactly $H D(S)$, where $D(S)=\max_{i<j}\norm{s_i-s_j}_2$ is the Euclidean row diameter. This diameter replaces the spectral norm in the resulting Muon-style steepest descent problem.
An exact solution is possible, but its direct formulation contains one $d$-dimensional vector variable for every token pair. For a vocabulary size of approximately $50$k, this means more than one billion token pairs, making the calculation impractical at every training step. We instead impose a stronger common-ball constraint and derive projected RowNorm as an $O(Vd)$ solution.
For the exact RowNorm oracle, we prove that its first-order decrease is at least $1/\sqrt{2}$ of the exact diameter-constrained optimum.
With Muon on the backbone, experiments across three seeds at 190M, 380M, and 640M parameters show that RowNorm reduces mean final step diameters and empirical Hilbert RMS perturbations by factors of $45$--$60$ and $12$--$15$, respectively, with only a $0.0057$--$0.0153$ increase in mean final validation loss.
\end{abstract}

%\begin{keywords}%
%  List of keywords%
%\end{keywords}

\section{Introduction}

A language-model head (LM head) with weights $U\in\R^{V\times d}$ maps a hidden state $h\in\R^d$ to the probability distribution $p_U(h)=\softmax(Uh)$ over $V$ tokens. With the inception of Muon, recent work has shifted attention from coordinate-wise optimization toward updates that respect the geometry and function of each parameter block \citep{bernstein2024old, bernstein2024modular, large2024scalable}.
For an additive update to a hidden linear layer, the spectral norm measures the largest activation change that the update can produce on an input of unit norm.
Muon constructs its update using this spectral geometry \citep{liu2025muon, jordan2024muon, pethick2025trainingdeeplearningmodels}.

The spectral norm, however, does not faithfully measure the function of the LM head.
Its logits are immediately passed through softmax, so the relevant mathematical object is a probability distribution rather than a Euclidean activation.
Softmax removes any shift shared by all logits, whereas the spectral norm does not. To see this, let $a\in\R^d$ and consider the update
$S=\one a^\top\in\R^{V\times d}$, whose rows are all identical. For every hidden state $h$,
\[
p_{U+S}(h)
=
\softmax\!\left(Uh+(a^\top h)\one\right)
=
p_U(h).
\]
Thus, $U$ and $U+S$ produce exactly the same probability distribution for every input. Under softmax cross entropy, the update also changes neither the loss nor the gradient passed to the backbone.
Nevertheless, $\norm{S}_{\mathrm{op}}=\sqrt{V}\norm{a}_2$ can be arbitrarily large. A trust region intended to measure functional change should assign zero size to an update that the complete module removes exactly.
The spectral norm fails this requirement for the composed LM head and softmax module.
This distinction is consistent with current practice as standard Muon recipes typically leave the LM head in the care of AdamW \citep{loshchilov2018decoupled,jordan2024muon}.

Each output coordinate of the LM head corresponds to a particular vocabulary token.
\citet{lau2026symmetry} formalize this token structure together with softmax's invariance to shared logit shifts and derive an update specialized to the LM head.
These observations motivate the central question of our paper:
\begin{center}
    \emph{Should the size of an LM head update be measured by its effect on
    the logits, or by its effect on the probability distribution after
    softmax?}
\end{center}
We argue for the latter. We seek a distance on probability
distributions whose induced update geometry ignores shared logit shifts.
Hilbert's projective distance \citep{cohen2023hyperbolic}, formally defined in Section~\ref{sec:hilbert}, is a natural choice.
It measures the largest change in pairwise log odds, and shared logit shifts cancel from every such comparison. For an additive update $S$ with rows $s_i^\top$, we show that

\[
\sup_{\norm{h}_2\le H} \mathcal{H}\!\left(p_{U+S}(h), p_U(h)\right) = H\max_{i<j}\norm{s_i - s_j}_2
\]
Writing $D(S)=\max_{i<j}\norm{s_i-s_j}_2$, the identity above says that the functional size of an LM head update is the Euclidean diameter of its rows.
Let $G=\nabla_U\mathcal L$. For softmax cross entropy, $\one^\top G=0$. For a small update $S$, $\mathcal L(U+S) = \mathcal L(U)+\inner{G}{S}_F+o(\norm{S}_F)$,
where $\inner{G}{S}_F=\operatorname{tr}(G^\top S)$. For a diameter bound $\eta$, the corresponding steepest descent problem is therefore
\[
S^\star\in
\arg\min_{D(S)\leq\eta}
\inner{G}{S}_F.
\]
This has the same linear minimization form as the problem underlying Muon.
Muon constrains the spectral norm of the update; here, Hilbert distance and softmax induce the row diameter constraint.
The vector-valued Kantorovich--Rubinstein duality gives an exact vector-flow characterization of the optimal value
\citep{ciosmak2021optimaltransportvectormeasures, robertson2025generalizationwassersteindistancebeckmann}.
Appendix~\ref{app:diameter-duality} specializes this result to the LM head and gives a complete proof.

Unfortunately, the direct flow formulation introduces one $d$-dimensional variable for each of the $\binom{V}{2}$ token pairs, making it unsuitable for use at every training step. We therefore solve a more restrictive problem by requiring all update rows to lie in a common ball:
\[
\min_{S,\,\zeta\in\R^d}
\inner{G}{S}_F
\qquad\text{subject to}\qquad
\max_i\norm{s_i-\zeta}_2\leq\frac{\eta}{2}.
\]
Any two rows in this ball are at most $\eta$ apart, so every feasible update also satisfies the original diameter bound.
One centered solution of this problem is projected RowNorm, the same update derived from softmax symmetries by \citet{lau2026symmetry}.
It can be computed in $O(Vd)$ time without constructing token pairs.
Equation~\ref{eq:rownorm-guarantee} shows that projected RowNorm achieves at least $1/\sqrt{2}$ of the optimal first-order decrease under the original diameter constraint. The proof is provided in Appendix~\ref{app:rownorm-proof}.

\begin{figure*}[!ht]
\centering
\includegraphics[width=.32\textwidth]{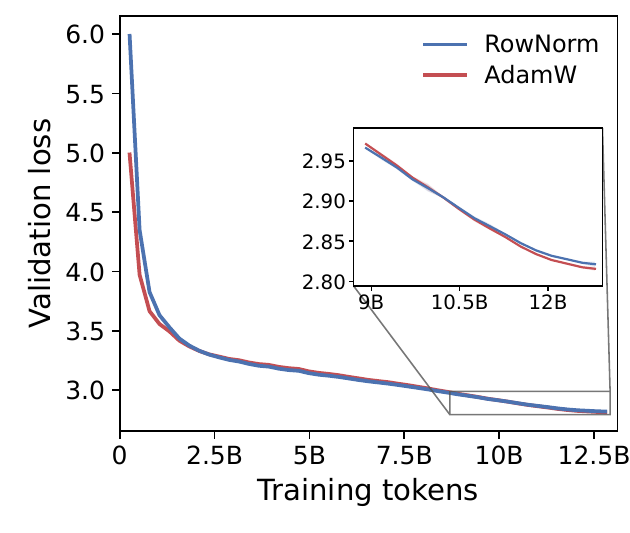}\hfill
\includegraphics[width=.32\textwidth]{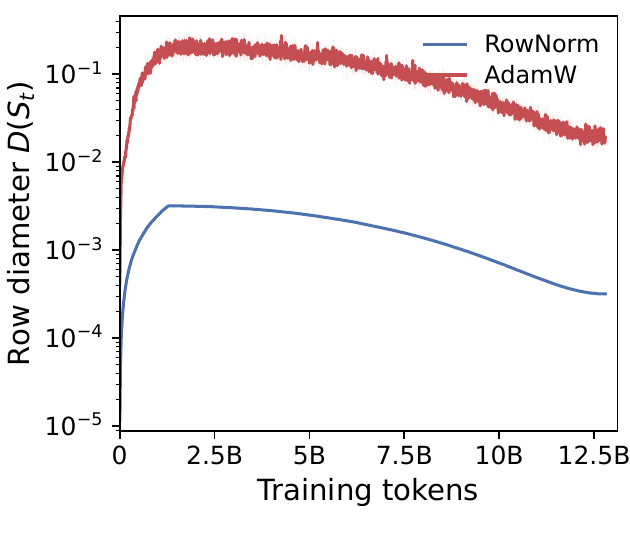}\hfill
\includegraphics[width=.32\textwidth]{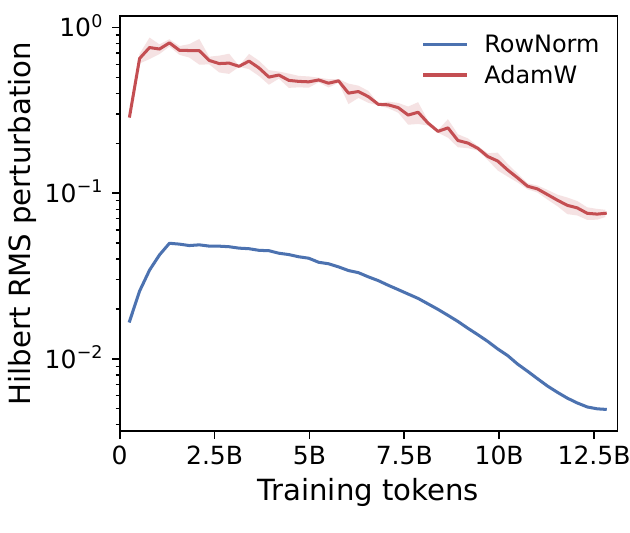}
\caption{\textbf{RowNorm vs AdamW at 640M parameters.}
Replacing the AdamW LM head recipe with RowNorm increases final validation loss by $0.0057$, while reducing the final row diameter and empirical per step Hilbert RMS perturbation by factors of $59.5$ and $15.2$, respectively.}
\label{fig:overview-640m}
\end{figure*}

Throughout, we study an untied LM head with no bias and softmax cross entropy, assuming that no other loss term depends on its weights.
Across 190M, 380M, and 640M models, RowNorm's mean final validation loss is at most $0.0153$ higher than AdamW's, while its optimizer increments have substantially smaller step diameters and empirical per step Hilbert perturbations. Figure~\ref{fig:overview-640m} previews the empirical comparison at our largest model scale. Complete experiments and details are provided in Appendix~\ref{app:experiments}. Our code is available at \url{https://github.com/AditSom-IFM/Hilbert-RowNorm}.

\section{Hilbert distance}
\label{sec:hilbert}

Softmax maps into the interior of the probability simplex
$\Delta_{V-1}^{\circ}
=\{p\in\R^V:p_i>0,\ \sum_i p_i=1\}$
\citep{birkhoff1957extensions,de1993hilbert}. For two distributions $p,q\in\Delta_{V-1}^{\circ}$, Hilbert's projective distance is
\begin{equation}
\mathcal{H}(p,q)=\max_i\log\frac{p_i}{q_i}-\min_i\log\frac{p_i}{q_i}.
\label{eq:hilbert-definition}
\end{equation}

To interpret equation~\ref{eq:hilbert-definition}, let \(a_i=\log\frac{p_i}{q_i}\). The difference between two coordinates is \(a_i-a_j=\log\frac{p_i}{q_i}-\log\frac{p_j}{q_j}=\log\frac{p_i}{p_j}-\log\frac{q_i}{q_j}\). Since the distance between the largest and smallest coordinates is the largest absolute difference between any two coordinates, \(\max_i a_i-\min_i a_i=\max_{i,j}\abs{a_i-a_j}\). Therefore,
\begin{equation}
\boxed{\mathcal{H}(p,q)=\max_{i,j}\abs{\log\frac{p_i}{p_j}-\log\frac{q_i}{q_j}}}
\label{eq:hilbert-log-odds}
\end{equation}
The quantity \(\log\frac{p_i}{p_j}\) is the log probability ratio of token $i$ to token $j$. Hilbert distance is therefore the largest change in pairwise log odds between $p$ and $q$. In particular, if \(\mathcal{H}(p,q)\le\varepsilon\), then every token pair satisfies
\(
e^{-\varepsilon}\leq\frac{p_i/p_j}{q_i/q_j}\leq e^{\varepsilon}.
\)
Thus, an upper bound on the Hilbert distance gives a simultaneous multiplicative bound on the change of every pairwise probability ratio. This property makes Hilbert distance useful for the LM head. It measures relative changes between tokens, rather than the absolute movement of individual probability coordinates.

\paragraph{Softmax and Hilbert distance.}
\label{sec:softmax}

Let \(p=\softmax(z)\) and \(q=\softmax(z')\) for two logit vectors $z,z'\in\R^V$. Softmax satisfies \(\frac{p_i}{p_j}=\frac{\exp(z_i)}{\exp(z_j)}=\exp(z_i-z_j)\), and therefore \(\log\frac{p_i}{p_j}=z_i-z_j\). Similarly, \(\log\frac{q_i}{q_j}=z_i'-z_j'\). Substituting these expressions into the pairwise form of Hilbert distance gives \(\mathcal{H}(p,q)=\max_{i,j}\abs{(z_i-z_j)-(z_i'-z_j')}=\max_{i,j}\abs{(z_i-z_i')-(z_j-z_j')}\). For a vector $x\in\R^V$, define its variation seminorm\footnote{This is a variation seminorm on $\R^V$ because it assigns zero size to every constant vector. It becomes a norm after logit vectors that differ by a common constant are identified.} by \(\varnorm{x}=\max_i x_i-\min_i x_i\). Since \(\max_{i,j}\abs{x_i-x_j}=\max_i x_i-\min_i x_i\), equation~\ref{eq:hilbert-log-odds} becomes
\begin{equation}
\boxed{\mathcal{H}\!\left(\softmax(z),\softmax(z')\right)=\varnorm{z-z'}}
\label{eq:softmax-Hilbert}
\end{equation}

The Hilbert distance between two softmax distributions therefore depends only on the difference between their logits \citep{de1993hilbert,nielsen2023non}.
The variation norm also ignores common logit shifts \(\varnorm{x+c\one}=\varnorm{x}\). This agrees with the corresponding softmax invariance \(\softmax(z+c\one)=\softmax(z)\). For an LM head update $S$, the logit vectors will be \(z=Uh\) and \(z'=(U+S)h\). Their difference is simply \(z'-z=Sh\).

\paragraph{Why choose Hilbert distance?}

Hilbert distance surprisingly connects with the Cross Entropy (CE) loss in addition to the softmax function. For a probability vector $p$, let $\mathcal{L}_c(p) = -\log p_c$ denote the CE loss when token $c$ is the target. For two distributions $p$ and $q$, the change in loss is \(\Delta\ell_c \coloneqq \mathcal{L}_c(p) - \mathcal{L}_c(q) = -\log p_c + \log q_c = \log \frac{q_c}{p_c}\). We can rewrite Hilbert distance to show \(\mathcal{H}(p, q) = \max_{i, j} \left|\log \frac{p_i}{q_i} - \log \frac{p_j}{q_j} \right| = \max_{i, j} \left| \Delta\ell_i - \Delta\ell_j \right|\). This leads us directly to 
\[
\mathcal{H}(p, q) = \max_c \Delta\ell_c - \min_c \Delta\ell_c = \max_c\left( \mathcal{L}_c(p) - \mathcal{L}_c(q) \right) - \min_c\left( \mathcal{L}_c(p) - \mathcal{L}_c(q) \right)
\]

Hilbert distance can be thought of as the largest difference between the CE loss changes of two possible target tokens. Moreover, since $p$ and $q$ are probability vectors summing to $1$, there must exist some $c, c'$ for which $q_c/p_c \geq 1$ and $q_{c'}/p_{c'} \leq 1$. This implies \(\min_c \Delta\ell_c \leq 0 \leq \max_c \Delta\ell_c\). It follows that for every target $c$,
\[
\boxed{
\left|\mathcal{L}_c(p) - \mathcal{L}_c(q)\right| \leq \mathcal{H}(p, q)
}
\]

Controlling Hilbert distance simultaneously bounds the change in CE loss for every possible target token.

\section{Geometry of an LM Head Update}
\label{sec:head-geometry}

\paragraph{Forward probability change.}

Let $S\in\R^{V\times d}$ have rows $s_i^\top$, and define its row diameter by $D(S)=\max_{i<j}\norm{s_i-s_j}_2$. For a hidden state $h$, the logit change is $Sh$, so equation~\ref{eq:softmax-Hilbert} gives $\mathcal H(p_{U+S}(h),p_U(h)) =\max_{i<j}\abs{(s_i-s_j)^\top h}$. Because the maximum is over finitely many token pairs, Euclidean duality gives
\begin{equation}
\boxed{
\sup_{\norm{h}_2\leq H}
\mathcal H\!\left(p_{U+S}(h),p_U(h)\right)
=
\max_{i<j}\sup_{\norm{h}_2\leq H}
\abs{(s_i-s_j)^\top h}
=
H D(S).
}
\label{eq:lm-head-diameter}
\end{equation}
Equation~\ref{eq:lm-head-diameter} shows that, for $H>0$, bounding the Hilbert distance by $\varepsilon$ uniformly over $\norm{h}_2\leq H$ is equivalent to $D(S)\leq\varepsilon/H$. The result depends on $S$, not on the current head $U$.

\paragraph{Common row shifts.}
Replacing $S$ by $S+\one a^\top$ adds the same scalar to every logit and changes neither the softmax distribution nor $D(S)$. Hence $D$ is a seminorm whose kernel consists of common-row matrices. We therefore choose the centered representative $\one^\top S=0$.

\paragraph{Backward gradient change.}
\label{sec:backward-change}

Let $r=\nabla_z\mathcal L$ be a fixed logit gradient. Since the loss depends on $z$ through softmax, $\one^\top r=0$. An update $S$ changes the gradient propagated to the hidden state by $S^\top r$, with
\begin{equation}
\boxed{\norm{S^\top r}_2\leq\frac{D(S)}{2}\norm{r}_1.}
\label{eq:fixed-cotangent-backward-bound}
\end{equation}
For cross entropy with target token $c$, $r=p-e_c$ and $\norm{r}_1/2=1-p_c$, giving $\norm{S^\top(p-e_c)}_2\leq D(S)(1-p_c)\leq D(S)$.  Appendix~\ref{app:lm-head-backward-change} proves this bound and treats the case where $S$ also changes the logits and $r$\footnote{Appendix~\ref{app:muon-backward-change} gives the spectral analogue.}.

\section{Steepest Descent for the LM Head}
\label{sec:steepest-descent}

The diameter constraint does not distinguish $S$ from $S+\one a^\top$. Let $G=\nabla_U\mathcal L$, with rows $g_i^\top$. For softmax cross entropy, each token position contributes $(\hat y-y)h^\top$ to $G$, and hence $\one^\top G=0$. Define the row-centering projector $P=I_V-\one\one^\top/V$. For every update $S$,
\[
\one^\top PS=0,
\qquad
D(PS)=D(S),
\qquad
\inner{G}{PS}_F=\inner{G}{S}_F.
\]
Thus, centering chooses one representative without changing either the functional size or the linearized loss. The steepest descent problem may therefore be written as
\begin{equation}
\boxed{
S^\star\in
\arg\min_{\substack{D(S)\leq\eta\\\one^\top S=0}}
\inner{G}{S}_F.
}
\label{eq:diameter-steepest-descent}
\end{equation}

Prior work on finite vector-valued Kantorovich--Rubinstein and Beckmann duality gives an exact vector-flow characterization of equation~\ref{eq:diameter-steepest-descent} \citep{ciosmak2021optimaltransportvectormeasures, robertson2025generalizationwassersteindistancebeckmann}.
Appendix~\ref{app:diameter-duality} specializes the result to the LM head. It bounds the first-order decrease of every feasible update and proves that the bound is attained. The corresponding flow program contains one $d$-dimensional variable for every token pair, making it impractical to solve at each training step. We therefore seek an $O(Vd)$ approximation that uses no pairwise variables.

\subsection{Projected RowNorm as an approximation}
\label{sec:projected-rownorm}

In this section, we show that projected RowNorm provides such an approximation.
Projected RowNorm replaces the diameter constraint with a stronger radius constraint.
Define $\operatorname{rad}(S)=\min_{\zeta\in\R^d}\max_i\norm{s_i-\zeta}_2$.
The triangle inequality gives $D(S)\leq2\operatorname{rad}(S)$.
Therefore, $\operatorname{rad}(S)\leq\eta/2$ implies $D(S)\leq\eta$. For $A\in\R^{V\times d}$ with rows $a_i^\top$, let $\mathsf{RN}(A)$ have row $a_i^\top/\norm{a_i}_2$ when $a_i\neq0$, and row $0^\top$ otherwise.
Define
\begin{equation}
\boxed{
\begin{aligned}
\mathsf{PRN}(A)
&=P\,\mathsf{RN}(PA),
\qquad
S_{\mathrm{RN}}(G;\eta)
=-\frac{\eta}{2}\mathsf{PRN}(G),
\\
S_{\mathrm{RN}}(G;\eta)
&\in
\arg\min_{\substack{\operatorname{rad}(S)\leq\eta/2\\
                    \one^\top S=0}}
\inner{G}{S}_F,
\qquad
\Delta_{\mathrm{RN}}
\coloneqq
-\inner{G}{S_{\mathrm{RN}}(G;\eta)}_F
=
\frac{\eta}{2}\sum_i\norm{g_i}_2.
\end{aligned}
}
\label{eq:projected-rownorm}
\end{equation}
Indeed, the center of the enclosing ball does not affect the objective because $\sum_i g_i=0$.
Before imposing zero row mean, fixing the center makes the minimization separate across rows. Each nonzero displacement $s_i-\zeta$ then points opposite to $g_i$.
Translating this solution to zero row mean preserves its radius and objective, giving equation~\ref{eq:projected-rownorm}. Appendix~\ref{app:rownorm-proof} gives the complete proof.
\citet{lau2026symmetry} derive the same projected RowNorm map from softmax symmetries.
Independently, Table~2 of \citet{pethick2025trainingdeeplearningmodels} identifies row normalization, up to their width-scaling convention, as the linear minimization oracle for the maximum row-Euclidean norm.
Here, it arises as an exact solution of the radius-restricted problem induced by the Hilbert row diameter geometry.

\paragraph{How much does RowNorm lose?}
\label{sec:rownorm-comparison}

Let $\Delta_\star=-\inner{G}{S^\star}_F$. If $G\neq0$, $\eta>0$, and
$k=\rank(G)$, then
\begin{equation}
\boxed{
\sqrt{\frac{k+1}{2k}}
\leq
\frac{\Delta_{\mathrm{RN}}}{\Delta_\star}
\leq1.
}
\label{eq:rownorm-guarantee}
\end{equation}
Since projected RowNorm satisfies the original diameter constraint, $\Delta_{\mathrm{RN}}\leq\Delta_\star$. For the lower bound, project the update rows onto the $k$-dimensional span of the gradient rows. This preserves the objective and cannot increase the diameter. Jung's theorem \citep{Jung1901,article} places the projected rows inside a ball of radius $\eta\sqrt{k/[2(k+1)]}$. Rescaling this ball to radius $\eta/2$ gives the stated factor. Appendix~\ref{app:rownorm-proof} provides the complete proof and a sharp equality construction.

\paragraph{Practical projected RowNorm.}
\label{sec:rownorm-algorithm}

At iteration $t$, let $G_t=\nabla_U\mathcal L_t(U_t)$, update $B_t=\mu B_{t-1}+(1-\mu)G_t$ with $B_{-1}=0$, form the bias corrected moment $\widehat B_t=B_t/(1-\mu^{t+1})$, and set $C_t=P\widehat B_t$. For $\epsilon_{\mathrm{num}}>0$, define $\left[\mathsf{RN}_{\epsilon_{\mathrm{num}}}(C_t)\right]_{i:} = c_{t,i}^\top/(\norm{c_{t,i}}_2+\epsilon_{\mathrm{num}})$. The practical update is
\begin{equation}
\boxed{
S_t^{\mathrm{RN}}
=-\gamma_tP\,\mathsf{RN}_{\epsilon_{\mathrm{num}}}(C_t),
\qquad
U_{t+1}=(1-\gamma_t\lambda)U_t+S_t^{\mathrm{RN}}.
}
\label{eq:rownorm-final-algorithm}
\end{equation}
For the update in equation~\ref{eq:rownorm-final-algorithm}, every normalized row has norm at most one, and centering preserves pairwise differences. Therefore $D(S_t^{\mathrm{RN}})\leq2\gamma_t$, corresponding to $\eta_t=2\gamma_t$. The update costs $O(Vd)$, uses one first-moment buffer, and stores no token pairs.
Across 190M, 380M, and 640M models, the AdamW and RowNorm LM head recipes differ by at most $0.0153$ in final validation loss. RowNorm gives $45$--$60\times$ smaller final step diameters and $12$--$15\times$ smaller empirical per step Hilbert RMS perturbations. The experiment setup and training curves are provided in Appendix~\ref{app:experiments}.

\section{Conclusion}
Looking at the LM head from first principles and considering its composition with softmax leads us to Hilbert's projective distance.
We identify row diameter as the trust region induced by this metric for the LM head.
The finite vector-valued Kantorovich--Rubinstein duality gives an exact vector-flow characterization of the resulting steepest descent problem \citep{ciosmak2021optimaltransportvectormeasures, robertson2025generalizationwassersteindistancebeckmann}.
Its direct formulation, however, is computationally expensive at vocabulary scale.
We therefore turn to projected RowNorm as a tractable approximation. The update itself is not new \citep{lau2026symmetry, pethick2025trainingdeeplearningmodels}.
Our contribution is to derive projected RowNorm from the probability geometry of the LM head and prove its sharp approximation guarantee relative to the exact diameter-constrained steepest descent problem.
Experiments from 190M to 640M parameters show substantially smaller step diameters and Hilbert perturbations, while validation loss remains slightly higher than AdamW.
More broadly, these results suggest that optimizer geometry should reflect the function and invariances of the complete module, rather than those of its parameter matrix alone.

\bibliography{sample}
\newpage
\clearpage
\appendix

\section{Related work}
\label{app:related-work}

\paragraph{Norms, modules, and Muon.}
A steepest descent direction is defined only after choosing how the size of an
update is measured. Changing this norm changes the resulting optimizer. Early
work used Schatten-$\infty$ geometry to derive stochastic spectral descent for
restricted Boltzmann machines \citep{pmlr-v38-carlson15}. Muon applies the
same spectral viewpoint to hidden matrices by using Newton--Schulz iteration
to approximate the polar factor of a momentum matrix
\citep{jordan2024muon}. With exact orthogonalization, this direction is the
linear minimization oracle of a spectral norm ball
\citep{bernstein2024old,pethick2025trainingdeeplearningmodels}. Subsequent work
scaled Muon to large language models \citep{liu2025muon} and interpreted its
combination with decoupled weight decay as implicit spectral norm constrained
optimization \citep{chen2025muonoptimizesspectralnorm}. More broadly, modular
norm and modular duality \citep{modula-docs} assign a geometry to each parameter module and
compose these choices across the architecture
\citep{large2024scalable,bernstein2024modular}; the Modula project implements
this program \citep{modula-docs}. We follow the same modulewise principle, but
take the composed LM head and softmax as our starting point and derive its
update geometry from the change induced in the output probability
distribution.

\paragraph{The LM head and softmax.}
The LM head is a vocabulary-indexed linear map whose logits are converted by
softmax into the next token distribution. Prior work has studied the
consequences of this structure in both the forward and backward passes. In the
forward pass, the rank of the linear-softmax parameterization restricts the
distributions it can represent, producing the softmax bottleneck
\citep{yang2017breaking,chang2022softmax}. In the backward pass, the transpose
of the head compresses vocabulary-space gradients, although whether this
compression itself causes harmful optimization dynamics remains under
discussion \citep{godey2026lost,murugan2026doeslmheadcreate}. The softmax map
has separately been studied through its relation to log-sum-exp and its
monotonicity and Lipschitz properties
\citep{gao2017properties,nair2025softmax}. Closest to our optimizer question,
\citet{lau2026symmetry} derive projected RowNorm from permutation symmetry of
the token rows and invariance to a shared logit shift. We obtain the same
update from a complementary functional argument by measuring the change it
induces in the output probability distribution.

\paragraph{Geometry on probability distributions.}
The probability simplex admits several geometries, each encoding a different
notion of change. The Fisher metric
gives rise to natural gradient, while K-FAC provides a tractable layerwise
approximation \citep{amari1998natural,martens2015optimizing}. Aitchison
geometry instead takes a log ratio view, using Euclidean distance between
centered log ratio coordinates
\citep{10.1111/j.2517-6161.1982.tb01195.x}. Hilbert's projective metric is
also expressed through log ratios, but measures their range rather than their
Euclidean magnitude. It originates in the geometry of positive cones and
Birkhoff contraction theory \citep{birkhoff1957extensions,de1993hilbert}, and
on the probability simplex it becomes the variation-norm distance between
normalized log coordinates \citep{gaubert2015dobrushin, Nielsen_2018, nielsen2023non}. This geometry also admits
quantitative comparisons with total variation
\citep{cohen2023hyperbolic}. We apply Hilbert distance to the output of the
composed LM head and softmax and show that the resulting worst case
distributional change is governed exactly by the Euclidean diameter of the
LM head update rows.

\paragraph{Row and column normalization.}
Row and column normalization have emerged from several different views of optimizer geometry. From the operator norm perspective, \citet{pethick2025trainingdeeplearningmodels} derive ColNorm and RowNorm as the exact linear minimization oracles of the $\ell_1$-to-RMS and RMS-to-$\ell_\infty$ norm balls, building on general maximum-column and maximum-row operator norm identities \citep{bernstein2024old,bernstein2024modular}. \citet{xu2026widthscalingneuraloptimizers} extends this construction to mean-normalized $p\rightarrow q$ operator norms, producing generalized row and column normalization with scaling chosen for stable behavior across model widths. For the LM head, \citet{glentis2026memoryefficientllmpretrainingminimalist} apply column normalization under a $d\times V$ convention; since columns index tokens, this is unprojected RowNorm after transposing to our $V\times d$ convention. Closest to our work, \citet{lau2026symmetry} derive the projected map $P\,\mathsf{RN}(P\,\cdot)$ from token permutation and logit-shift symmetries. Row normalization has also been motivated as an efficient approximation to Muon-style preconditioning \citep{deng2026rmnprowmomentumnormalizedpreconditioning}, while gradient multi-normalization alternates row and column normalization to satisfy several norm constraints simultaneously \citep{scetbon2025gradientmultinormalizationstatelessscalable}. Although these methods produce related algebraic operations, they arise from different constraints. We instead begin with the composed LM head and softmax: Hilbert distance induces the translation invariant row diameter geometry, and projected RowNorm solves its covering-radius restriction with the approximation guarantee in equation~\ref{eq:rownorm-guarantee}.

\paragraph{Vector flows and optimal transport.}
Optimal transport admits complementary potential and flow formulations. In
the Beckmann formulation, transport minimizes the total magnitude of a flux
whose divergence matches the prescribed sources and sinks
\citep{b89ef164-61eb-3086-8dda-c0a165a1b822}. Kantorovich--Rubinstein duality
relates this flow problem to optimization over Lipschitz potentials, and this
correspondence extends from scalar to vector measures
\citep{ciosmak2021optimaltransportvectormeasures,
ciosmak2021matrixholdersinequalitydivergence}. On finite graphs,
\citet{robertson2025generalizationwassersteindistancebeckmann} establish the
corresponding duality between $\ell_{2,1}$ vector flows and vector-valued
potentials. Their formulation specializes to our minimum-tension problem on
the complete graph with unit edge weights and identity connection. We use
this flow--potential duality to characterize steepest descent under the
LM head row diameter geometry.

\section{Pretraining experiments}
\label{app:experiments}

In this section, we compare AdamW and RowNorm as optimizers for the LM head during pretraining. We track validation loss and, at fixed training intervals,
the row diameter and induced Hilbert perturbation of the derived LM
head increment. Within each model size and seed, the paired runs
share the backbone and input embedding optimizers, initialization seed, data
order, batch configuration, and training schedule. The experimental differences are isolated at the LM head.

\paragraph{Models and data.}
We train decoder-only Transformers \citep{vaswani2023attentionneed} with 32
pre-norm blocks and a final RMSNorm before the LM head \citep{zhang2019rootmeansquarelayer}.
Each block uses causal attention, SwiGLU MLPs of
width $4d$ \citep{shazeer2020gluvariantsimprovetransformer}, attention head dimension 64, query and key
RMS normalization, and RoPE with base $10{,}000$
\citep{su2023roformerenhancedtransformerrotary}.
All RMS normalizations use $\epsilon=10^{-6}$ and no
learned gain. The input embedding and LM head are untied. Linear maps have no
bias, dropout is zero, and all embedding and linear weights have i.i.d. entries
drawn from $\mathcal N(0,0.02^2)$. Table~\ref{tab:experiment-models} gives the
model dimensions and training scales. Parameter counts include both the input
embedding and the untied LM head.
We use 890 training shards and one validation shard from the FineWeb100B
release, tokenized with the GPT-2 tokenizer \citep{radford2019language, penedo2024finewebdatasetsdecantingweb}. They contain 89.0 billion
training tokens and 100 million validation tokens. The vocabulary contains
$50{,}257$ tokens and the context length is $2{,}048$. At
each loader epoch, the shard order and the order of complete optimizer batches
within each shard are shuffled deterministically.
Validation uses a fixed sequential prefix
of $5{,}242{,}880$ tokens and reports next token cross entropy.

\begin{table*}[t]
\centering
\small
\setlength{\tabcolsep}{5pt}
\begin{tabular}{lrrrrrrr}
\toprule
Model & $d$ & Heads & Parameters & Micro/GPU & Accum. & Updates & Training tokens \\
\midrule
190M & 512  & 8  & $185{,}680{,}896$ & 32 & 1 & $7{,}083$  & $3{,}713{,}531{,}904$ \\
380M & 768  & 12 & $379{,}184{,}640$ & 16 & 2 & $14{,}464$ & $7{,}583{,}301{,}632$ \\
640M & 1024 & 16 & $639{,}797{,}248$ & 8  & 4 & $24{,}406$ & $12{,}795{,}772{,}928$ \\
\bottomrule
\end{tabular}
\caption{\textbf{Model and training scales.} Microbatch is measured in sequences per GPU. All runs use eight GPUs, a global batch of 256 sequences, and 20 training tokens per parameter.}
\label{tab:experiment-models}
\end{table*}

\paragraph{Training protocol.}
For every model size and LM head recipe, we run seeds 0, 1, and 2, giving 18 runs in total, each model trained for 20 tokens per parameter. The global batch contains 256 sequences of length $2{,}048$, or $524{,}288$ tokens per update. For the 190M, 380M, and 640M models, the per-GPU microbatch sizes are 32, 16, and 8, with 1, 2, and 4 gradient accumulation steps, respectively. We clip the global gradient norm at 1.0. Let $T$ denote the total number of updates. Every learning rate follows the same schedule: linear warmup for $\lfloor 0.1T\rfloor$ updates, followed by cosine decay from its peak to $10\%$ of the peak at the final update. Validation is performed every 500 updates and at the final update. Training uses one node with eight NVIDIA H200 GPUs and distributed data parallelism. Parameters and optimizer states use FP32. Forward passes use BF16 autocast; RMSNorm computations use FP32 internally, and logits are cast to FP32 before cross entropy and diagnostic calculations. The runs use PyTorch~\texttt{2.13.0+cu126} with CUDA~12.6.

\paragraph{Optimizer recipes.}
We apply Nesterov Muon to every linear weight matrix in the Transformer
blocks. It uses peak learning rate $8\times10^{-3}$, momentum $0.95$,
decoupled weight decay $0.1$, five Newton--Schulz iterations, and
$\epsilon=10^{-5}$. For a matrix of shape $m\times n$, the orthogonalized
update is scaled by $\sqrt{\max\{1,m/n\}}$. The input embedding uses AdamW
with peak learning rate $4\times10^{-3}$,
$(\beta_1,\beta_2)=(0.9,0.999)$, $\epsilon=10^{-10}$, and weight decay $0.1$.
These backbone and input embedding settings are identical in every run. Only the LM head recipe changes. The AdamW head uses
$(\beta_1,\beta_2)=(0.9,0.999)$, $\epsilon=10^{-10}$, and weight decay $0.1$. For RowNorm, the moment is updated as
$B_t=0.95B_{t-1}+0.05G_t$, with $B_{-1}=0$, and stabilized row normalization
is applied to the bias-corrected moment $B_t/(1-0.95^{t+1})$. RowNorm uses
$\epsilon=10^{-8}$ and no LM head weight decay. Table~\ref{tab:experiment-head-optimizers} gives the peak learning rates.
The comparison changes the complete LM head recipe, including its
learning rate, moment estimate, numerical stabilizer, and weight decay.

\begin{table}[t]
\centering
\small
\begin{tabular}{lcc}
\toprule
Model & AdamW head LR & RowNorm head LR \\
\midrule
190M & $4.0\times10^{-3}$   & $3.2\times10^{-3}$ \\
380M & $2.667\times10^{-3}$ & $2.133\times10^{-3}$ \\
640M & $2.0\times10^{-3}$   & $1.6\times10^{-3}$ \\
\bottomrule
\end{tabular}
\caption{\textbf{Peak learning rates for the LM head}. RowNorm uses momentum
$\beta=0.95$ at every model scale.}
\label{tab:experiment-head-optimizers}
\end{table}

The implementation normalizes the EMA rows directly rather than explicitly
materializing both projections in $P\,\mathsf{RN}(P\,\cdot)$. In exact
arithmetic, the inner projection is redundant because $B_{-1}=0$ and
$\one^\top G_t=0$ imply $\one^\top B_t=0$. The outer projection only
subtracts a common row from the update. For the untied softmax head considered
here, this changes neither the output distribution, loss, row diameter,
Hilbert perturbation, nor the gradient passed to the backbone. The implemented
update therefore represents the same quotient-space direction as projected
RowNorm, up to floating-point roundoff.

\paragraph{Diagnostics.}
We compute all geometric diagnostics from the optimizer increment, excluding
decoupled weight decay. Let $U_t$ be the LM head before update $t$, and let
$S_t$ be the signed optimizer increment formed from the globally clipped
gradient and optimizer state, including the current learning rate. The complete
update is
\[
U_{t+1}=(1-\gamma_t\lambda)U_t+S_t,
\]
where $\lambda=0.1$ for AdamW and $\lambda=0$ for RowNorm. We record the row diameter
\[
D(S_t)=\max_{i<j}\norm{(S_t)_{i:}-(S_t)_{j:}}_2
\]
every 10 updates and at the final update. For the Hilbert diagnostic, we fix a panel of $N=8{,}192$ validation token positions. After update $t$, we recompute the hidden states of this panel under the updated model and denote them by $h_{t,\ell}^{\mathrm{post}}$. The perturbation at position
$\ell$ is
\[
\delta_{t,\ell}
=
\max_i(S_t h_{t,\ell}^\mathrm{post})_i-\min_i(S_t h_{t,\ell}^\mathrm{post})_i.
\]
We report the root mean square
\[
\left(\frac{1}{N}\sum_{\ell=1}^N \delta_{t,\ell}^2\right)^{1/2}
\]
every 500 updates and at the final update. This statistic measures a single
optimizer increment on the fixed validation panel. It is not the supremum over
all bounded hidden states or the cumulative drift from initialization.

\begin{figure*}[t]
\centering
\includegraphics[width=.32\textwidth]{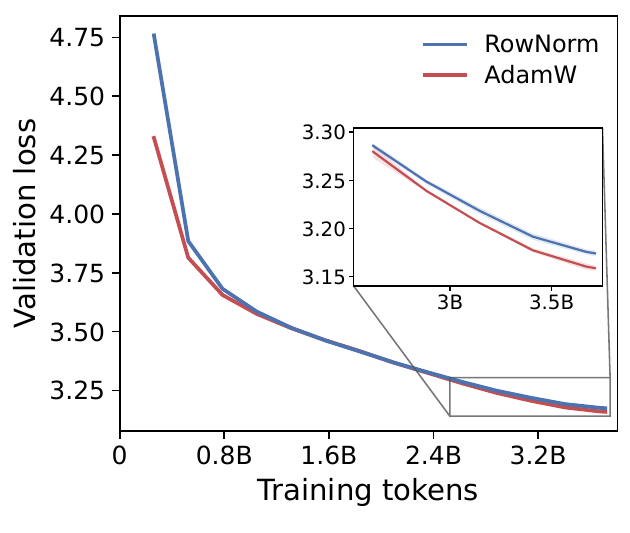}\hfill
\includegraphics[width=.32\textwidth]{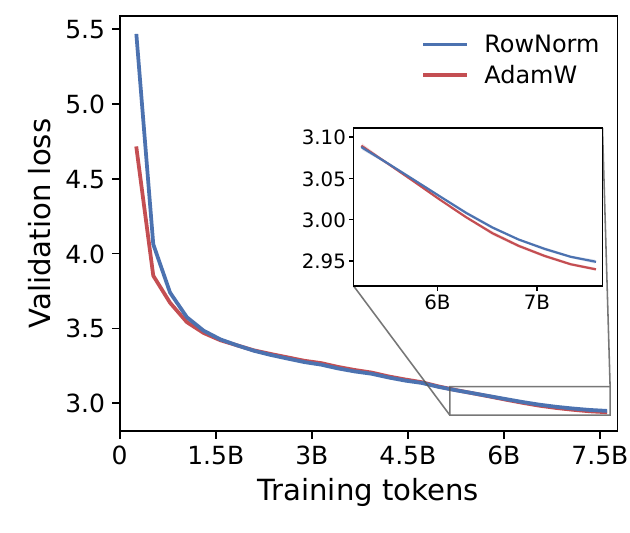}\hfill
\includegraphics[width=.32\textwidth]{Plots/640M/loss_vs_training_tokens.pdf}
\caption{\textbf{Validation cross entropy} for the 190M, 380M, and 640M models,
respectively. Solid curves show the pointwise mean over seeds 0, 1, and 2, while
shaded bands show the mean $\pm$ one sample standard deviation. The insets enlarge the final portion of training.}
\label{fig:experiment-loss}
\end{figure*}

\begin{figure*}[t]
\centering
\includegraphics[width=.32\textwidth]{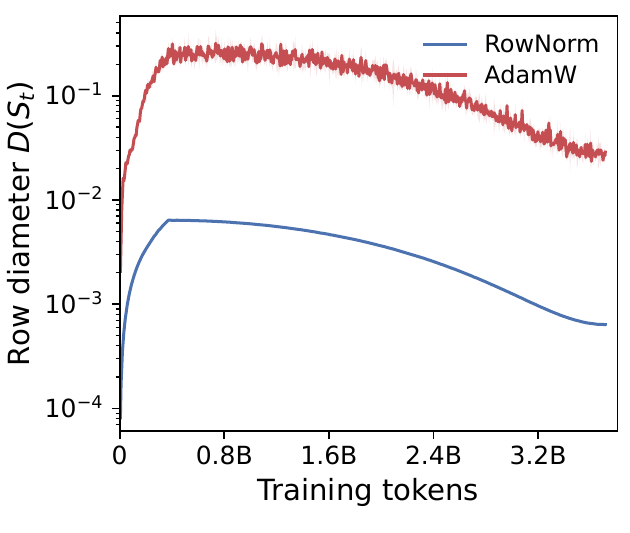}\hfill
\includegraphics[width=.32\textwidth]{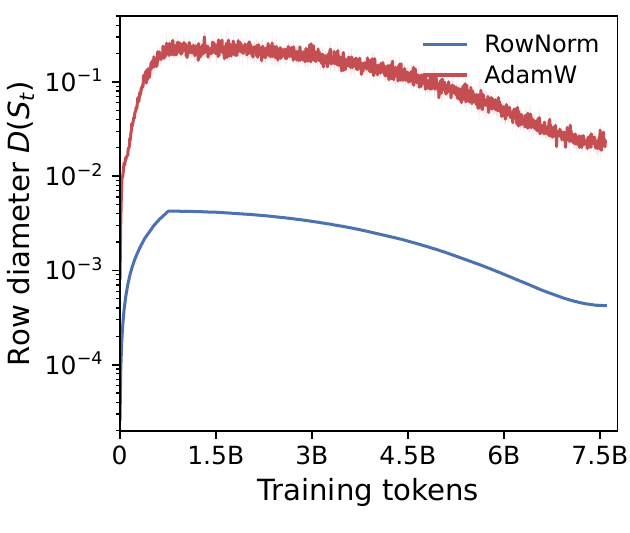}\hfill
\includegraphics[width=.32\textwidth]{Plots/640M/diameter_vs_training_tokens.pdf}
\caption{\textbf{Row diameter of the LM head step.} Solid curves show the
pointwise mean over seeds 0, 1, and 2, while shaded bands show the mean $\pm$ one
sample standard deviation. For every seed and model size, RowNorm produces a substantially smaller and stable diameter throughout training. The vertical axis is logarithmic.}
\label{fig:experiment-diameter}
\end{figure*}

\begin{figure*}[t]
\centering
\includegraphics[width=.32\textwidth]{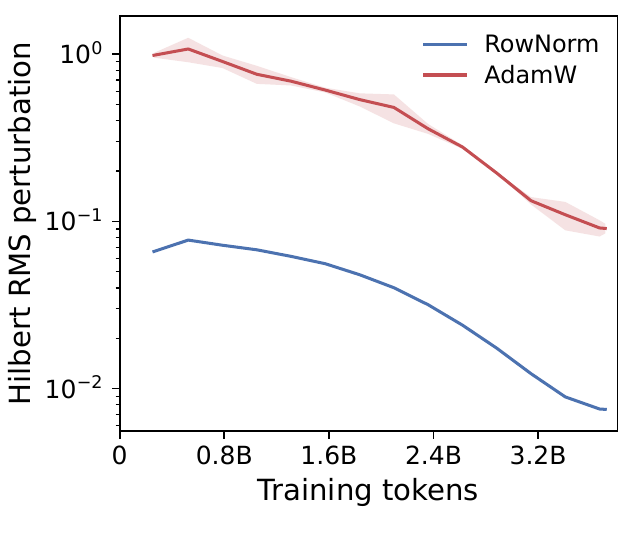}\hfill
\includegraphics[width=.32\textwidth]{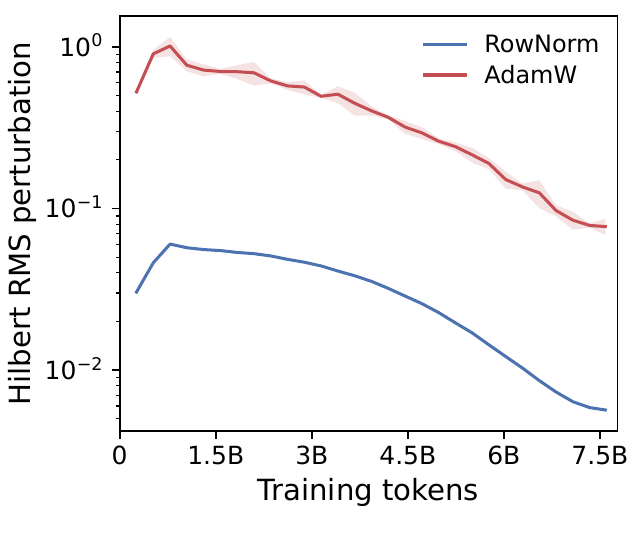}\hfill
\includegraphics[width=.32\textwidth]{Plots/640M/hilbert_perturbation_vs_training_tokens.pdf}
\caption{\textbf{Empirical per step Hilbert RMS perturbation.} Solid curves
show the pointwise mean over seeds 0, 1, and 2, while shaded bands show the mean
$\pm$ one sample standard deviation. For every seed and model size, RowNorm is below AdamW at every recorded evaluation. The vertical axis is logarithmic.}
\label{fig:experiment-hilbert}
\end{figure*}

\paragraph{Results.}
Table~\ref{tab:experiment-results} summarizes the final validation losses and geometric diagnostics.
Across the three paired seeds, the final validation loss differences
(RowNorm minus AdamW) are $0.0153\pm0.0027$, $0.0091\pm0.0007$, and
$0.0057\pm0.0012$ at 190M, 380M, and 640M parameters, respectively, where
$\pm$ denotes one sample standard deviation across seeds
(Figure~\ref{fig:experiment-loss}). At the final update, the ratios of the
AdamW mean to the RowNorm mean are $45.1$, $54.3$, and $59.5$ for row
diameter (Figure~\ref{fig:experiment-diameter}) and $12.1$, $13.6$, and
$15.2$ for Hilbert RMS (Figure~\ref{fig:experiment-hilbert}). For every
model size and seed, RowNorm has smaller row diameter and Hilbert RMS at every
recorded checkpoint.

\begin{table*}[t]
\centering
\small
\setlength{\tabcolsep}{4pt}
\begin{tabular}{llccc}
\toprule
Model & LM head & Validation loss & $D(S_t)$ & Hilbert RMS \\
\midrule
190M & AdamW
 & $3.1588\pm0.0027$
 & $(2.870\pm0.140)\times10^{-2}$
 & $(9.076\pm0.605)\times10^{-2}$ \\
     & RowNorm
 & $3.1741\pm0.0031$
 & $(6.369\pm0.009)\times10^{-4}$
 & $(7.506\pm0.017)\times10^{-3}$ \\
\addlinespace
380M & AdamW
 & $2.9403\pm0.0019$
 & $(2.303\pm0.324)\times10^{-2}$
 & $(7.732\pm0.874)\times10^{-2}$ \\
     & RowNorm
 & $2.9494\pm0.0024$
 & $(4.241\pm0.015)\times10^{-4}$
 & $(5.667\pm0.024)\times10^{-3}$ \\
\addlinespace
640M & AdamW
 & $2.8156\pm0.0024$
 & $(1.895\pm0.368)\times10^{-2}$
 & $(7.555\pm0.382)\times10^{-2}$ \\
     & RowNorm
 & $2.8213\pm0.0033$
 & $(3.1856\pm0.0029)\times10^{-4}$
 & $(4.955\pm0.025)\times10^{-3}$ \\
\bottomrule
\end{tabular}
\caption{\textbf{Final validation loss and LM head geometry.} Entries are
means $\pm$ sample standard deviations over seeds 0, 1, and 2. Geometry is
computed from $S_t$ at the final update and excludes decoupled weight decay.}
\label{tab:experiment-results}
\end{table*}

\FloatBarrier

\section{Spectral control of the forward and backward maps}
\label{app:muon-backward-change}

In this section, we show why the spectral norm controls the change in both
maps of a linear layer. Let $y=Wx$, where $W\in\R^{m\times n}$, and let
$S\in\R^{m\times n}$ be an update to its weights. For a fixed input
$x\in\R^n$, the change in the forward map is
\[
(W+S)x-Wx=Sx,
\qquad
\norm{Sx}_2\leq\norm{S}_{\mathrm{op}}\norm{x}_2.
\]
Consequently, for any $H\geq0$,
\[
\sup_{\norm{x}_2\leq H}
\norm{(W+S)x-Wx}_2
=
H\norm{S}_{\mathrm{op}}.
\]

Now fix an output gradient $g_y\in\R^m$. Before the update, the layer
propagates this gradient as $W^\top g_y$; afterward, it propagates it as
$(W+S)^\top g_y$. Their difference satisfies
\[
(W+S)^\top g_y-W^\top g_y=S^\top g_y,
\qquad
\norm{S^\top g_y}_2
\leq
\norm{S^\top}_{\mathrm{op}}\norm{g_y}_2
=
\norm{S}_{\mathrm{op}}\norm{g_y}_2.
\]
Thus, for any $R\geq0$,
\[
\sup_{\norm{g_y}_2\leq R}
\norm{(W+S)^\top g_y-W^\top g_y}_2
=
R\norm{S}_{\mathrm{op}}.
\]
The identity
$\norm{S^\top}_{\mathrm{op}}=\norm{S}_{\mathrm{op}}$ therefore makes the
same spectral bound control the worst case change in both the forward map
and its transpose.

\section{Backward change through the LM head and softmax}
\label{app:lm-head-backward-change}

In this section, we bound the complete change in the gradient propagated
through the LM head. The calculation in
equation~\ref{eq:fixed-cotangent-backward-bound} holds the logit gradient
fixed. Here, the head update is also allowed to change the logits and hence
the softmax probabilities.

\paragraph{A fixed zero-sum logit gradient.}
We first prove the bound used in the main text. Let
$A\in\R^{V\times d}$ have rows $a_i^\top$, and let
$r\in\R^V$ satisfy $\one^\top r=0$. The claim is immediate when $r=0$, so
suppose $r\neq0$. Set $m=\norm{r}_1/2$ and write
$r=m(\pi^+-\pi^-)$, where
$\pi^+_i=(r_i)_+/m$ and $\pi^-_i=(-r_i)_+/m$. The zero-sum condition
implies that $\pi^+$ and $\pi^-$ are probability vectors. Therefore,
\[
A^\top r
=
m\left(\sum_i\pi^+_i a_i-\sum_j\pi^-_j a_j\right)
=
m\sum_{i,j}\pi^+_i\pi^-_j(a_i-a_j),
\]
and the triangle inequality gives
\begin{equation}
\norm{A^\top r}_2
\leq
m\sum_{i,j}\pi^+_i\pi^-_j\norm{a_i-a_j}_2
\leq
mD(A)
=
\frac{D(A)}{2}\norm{r}_1.
\label{eq:zero-sum-diameter-bound}
\end{equation}
Taking $A=S$ proves
equation~\ref{eq:fixed-cotangent-backward-bound}. The constant is exact:
choosing $r=e_i-e_j$ for a pair attaining the row diameter gives
\[
D(A)
=
\sup_{\substack{\one^\top r=0\\\norm{r}_1\leq2}}
\norm{A^\top r}_2.
\]

\paragraph{The complete backward change.}
Fix a hidden state $h$ and target token $c$, and define
\[
p=\softmax(Uh),
\qquad
q=\softmax((U+S)h),
\qquad
\delta=\mathcal H(q,p)=\varnorm{Sh}.
\]
The gradients propagated to $h$ before and after the update are
$b=U^\top(p-e_c)$ and $b^+=(U+S)^\top(q-e_c)$. Their difference has two equivalent decompositions:
\begin{equation}
\begin{aligned}
b^+-b
&=S^\top(q-e_c)+U^\top(q-p)\\
&=S^\top(p-e_c)+(U+S)^\top(q-p).
\end{aligned}
\label{eq:full-backward-decomposition}
\end{equation}
The first term in either line changes the transpose map while holding a logit gradient fixed. The second term accounts for the change in the softmax probabilities. All three vectors $q-e_c$, $p-e_c$, and $q-p$ have zero sum. Moreover,
\[
\frac{1}{2}\norm{q-e_c}_1=1-q_c,
\qquad
\frac{1}{2}\norm{p-e_c}_1=1-p_c,
\qquad
\frac{1}{2}\norm{q-p}_1=\operatorname{TV}(p,q).
\]
Applying equation~\ref{eq:zero-sum-diameter-bound} to equation~\ref{eq:full-backward-decomposition} therefore reduces the problem to bounding $\operatorname{TV}(p,q)$.

\paragraph{Total variation and Hilbert distance.}
For strictly positive probability vectors, the sharp comparison \citep[Proposition~7]{Reeb_2011}
(see also \citep[Theorem~5.1]{cohen2023hyperbolic}) is
\begin{equation}
\operatorname{TV}(p,q)
\leq
\tanh\!\left(\frac{\mathcal H(q,p)}{4}\right).
\label{eq:tv-hilbert-bound}
\end{equation}
To prove this, set $\rho_i=q_i/p_i$,
$a=\min_i\rho_i$, and $b=\max_i\rho_i$. Since $\sum_i p_i\rho_i=1$, we have $a\leq1\leq b$. The case $a=b=1$ is immediate. Otherwise, convexity of $\rho\mapsto\abs{\rho-1}$ gives, for
$\rho\in[a,b]$,
\[
\abs{\rho-1}
\leq
\frac{b-\rho}{b-a}(1-a)
+
\frac{\rho-a}{b-a}(b-1).
\]
Averaging with respect to $p$ and using
$\sum_i p_i\rho_i=1$ yields
\begin{equation}
\operatorname{TV}(p,q)
=
\frac{1}{2}\sum_i p_i\abs{\rho_i-1}
\leq
\frac{(b-1)(1-a)}{b-a}.
\label{eq:tv-chord-bound}
\end{equation}
The right-hand side of equation~\ref{eq:tv-chord-bound} can be bounded by
writing $R=b/a=e^\delta$ and $t=\sqrt R$, where
$\delta=\mathcal H(q,p)$. Then
\[
\frac{(b-1)(1-a)}{b-a}
=
\frac{(t^2a-1)(1-a)}{a(t^2-1)}
\leq
\frac{t-1}{t+1}.
\]
After multiplication by the positive denominator $a(t^2-1)$, the final inequality is exactly $(ta-1)^2\geq0$. Finally,
\[
\frac{t-1}{t+1}
=
\frac{e^{\delta/2}-1}{e^{\delta/2}+1}
=
\tanh\!\left(\frac{\delta}{4}\right),
\]
which proves equation~\ref{eq:tv-hilbert-bound}. The constant is sharp: the
two inequalities become equalities for a two-point likelihood ratio taking
the values $e^{-\delta/2}$ and $e^{\delta/2}$ with the appropriate masses. Combining equations~\ref{eq:zero-sum-diameter-bound},
\ref{eq:full-backward-decomposition}, and
\ref{eq:tv-hilbert-bound} gives
\begin{equation}
\boxed{
\begin{aligned}
\norm{b^+-b}_2
\leq\min\Bigg\{&
D(S)(1-q_c)+D(U)\tanh\!\left(\frac{\delta}{4}\right),\\
&
D(S)(1-p_c)+D(U+S)\tanh\!\left(\frac{\delta}{4}\right)
\Bigg\}.
\end{aligned}
}
\label{eq:full-backward-bound}
\end{equation}

\paragraph{A uniform diameter bound.}
Since
\[
\delta
=
\varnorm{Sh}
=
\max_{i,j}\abs{(s_i-s_j)^\top h}
\leq
D(S)\norm{h}_2,
\]
equation~\ref{eq:full-backward-bound} implies
\begin{equation}
\norm{b^+-b}_2
\leq
D(S)+D_{\min}\tanh\!\left(\frac{D(S)\norm{h}_2}{4}\right),
\qquad
D_{\min}=\min\{D(U),D(U+S)\}.
\label{eq:full-backward-diameter-bound}
\end{equation}
In particular, if $\norm{h}_2\leq H$ and $D(S)\leq\eta$, then
\[
\norm{b^+-b}_2
\leq
\eta+D_{\min}\tanh\!\left(\frac{H\eta}{4}\right)
\leq
\eta\left(1+\frac{H D_{\min}}{4}\right).
\]

The dependence on the current head in
equation~\ref{eq:full-backward-diameter-bound} is unavoidable. For example, fix
$\eta>0$ and take $V=d=2$, $h=e_1$, and rows
$u_1=Me_2$, $u_2=-Me_2$, $s_1=(\eta/2)e_1$, and
$s_2=-(\eta/2)e_1$. Then $D(S)=\eta$,
$U^\top(q-p)=M\tanh(\eta/2)e_2$, while
$S^\top(q-e_c)$ is parallel to $e_1$. Hence
$\norm{b^+-b}_2\geq M\tanh(\eta/2)\to\infty$ as $M\to\infty$.
Thus no bound on the complete backward change can depend on $D(S)$ alone.

\section{Exact dual characterization of the diameter problem}
\label{app:diameter-duality}

The diameter-constrained update problem has an equivalent representation on the complete graph whose vertices are tokens. The rows of the gradient specify an imbalance at each token, while vector-valued flows between token pairs balance these rows. A pairing identity first gives a lower bound on the update objective. The dual potential problem then shows that this bound is attained. Throughout, let $G\in\R^{V\times d}$ have rows $g_i^\top$, assume $\sum_i g_i=0$, and fix $\eta\geq0$. For $S\in\R^{V\times d}$, write
$s_i^\top$ for its $i$th row and recall that
\[
D(S)=\max_{i<j}\norm{s_i-s_j}_2.
\]

\paragraph{Balanced vector flows.}
For each pair $i<j$, let $y_{ij}\in\R^d$ denote the vector flow from token $i$ to token $j$, and extend it antisymmetrically $y_{ij} = -y_{ji}$. We say that a flow $Y=\{y_{ij}\}_{i<j}$ balances $G$ if
\[
\sum_{j\neq i}y_{ij}=g_i
\qquad\text{for every token }i.
\]
Thus, the net vector leaving token $i$ must equal the corresponding gradient
row. Among all balanced flows, define
\begin{equation}
\boxed{
\mathcal T(G)
\coloneqq
\min_{\substack{\{y_{ij}\in\R^d\}_{i<j}\\
                 \sum_{j\neq i}y_{ij}=g_i\ \text{for every }i}}
\sum_{i<j}\norm{y_{ij}}_2 .
}
\label{eq:minimum-tension}
\end{equation}
The objective is the total magnitude of the flow. We call its minimum $\mathcal T(G)$ the minimum total tension. The mechanical origin of this terminology is explained at the end of the section.

\paragraph{Exact characterization.}
The centered diameter oracle satisfies
\begin{equation}
\boxed{
\min_{\substack{D(S)\leq\eta\\\one^\top S=0}}
\inner{G}{S}_F
=
-\eta\mathcal T(G).
}
\label{eq:diameter-flow-duality}
\end{equation}
Equivalently, the largest attainable first-order decrease under the diameter
bound is $\eta\mathcal T(G)$. We now prove this identity.

\paragraph{Existence of a balanced flow.}
The condition $\sum_i g_i=0$ is necessary for balance, since antisymmetry
implies
\[
\sum_i\sum_{j\neq i}y_{ij}=0.
\]
On the complete graph, it is also sufficient. For each $i<j$, define
\[
y_{ij}^{(0)}=\frac{g_i-g_j}{V},
\qquad
y_{ji}^{(0)}=-y_{ij}^{(0)}.
\]
For every token $i$,
\[
\sum_{j\neq i}y_{ij}^{(0)}
=
\frac{1}{V}\sum_{j\neq i}(g_i-g_j)
=
\frac{(V-1)g_i-\sum_{j\neq i}g_j}{V}
=
g_i,
\]
Thus the feasible set in equation~\ref{eq:minimum-tension} is nonempty. The feasible set is a closed affine set. Moreover, every sublevel set of $\sum_{i<j}\norm{y_{ij}}_2$ is bounded, since the norm of each edge variable is bounded by the total sum. Its feasible sublevel sets are therefore compact, so the minimum defining $\mathcal T(G)$ is attained.

\paragraph{Pairing flows with updates.}
Balanced flows enter the update problem through the following identity. For
any balanced flow $Y$ and any update $S$,
\begin{equation}
\begin{aligned}
\inner{G}{S}_F
&=
\sum_i\inner{g_i}{s_i}
=
\sum_i
\inner{\sum_{j\neq i}y_{ij}}{s_i} \\
&=
\sum_{i<j}
\left(
\inner{y_{ij}}{s_i}
+
\inner{y_{ji}}{s_j}
\right)
=
\sum_{i<j}\inner{y_{ij}}{s_i-s_j}.
\end{aligned}
\label{eq:flow-pairing}
\end{equation}
The linear objective can therefore be written entirely in terms of pairwise
row differences. If $D(S)\leq\eta$, Cauchy--Schwarz gives
\[
\inner{G}{S}_F
\geq
-\sum_{i<j}
\norm{y_{ij}}_2\norm{s_i-s_j}_2
\geq
-\eta\sum_{i<j}\norm{y_{ij}}_2.
\]
Choosing a balanced flow that attains $\mathcal T(G)$ yields
\[
\inner{G}{S}_F\geq-\eta\mathcal T(G).
\]
Thus no feasible update can have objective value below
$-\eta\mathcal T(G)$. It remains to show that this lower bound is attained.

\paragraph{The potential dual.}
To determine whether the preceding lower bound is sharp, we derive the Lagrange dual of the minimum-flow problem. Associate a vector Lagrange multiplier $\phi_i\in\R^d$ with each balance equation, and let $\Phi\in\R^{V\times d}$ have rows $\phi_i^\top$.

\[
\begin{aligned}
\mathfrak L(Y,\Phi)
&=
\sum_{i<j}\norm{y_{ij}}_2
+
\sum_i
\inner{\phi_i}{
g_i-\sum_{j\neq i}y_{ij}
} \\
&=
\inner{G}{\Phi}_F
+
\sum_{i<j}
\left(
\norm{y_{ij}}_2
-
\inner{\phi_i-\phi_j}{y_{ij}}
\right),
\end{aligned}
\]
where the second equality uses $y_{ji}=-y_{ij}$. For any $a\in\R^d$,
\[
\inf_{y\in\R^d}
\left(
\norm{y}_2-\inner{a}{y}
\right)
=
\begin{cases}
0, & \norm{a}_2\leq1,\\
-\infty, & \norm{a}_2>1.
\end{cases}
\]
The first case follows from Cauchy--Schwarz. In the second case, taking
$y=t\,a/\norm{a}_2$ and sending $t\to\infty$ makes the expression tend to
$-\infty$. Applying this identity independently to every edge, the infimum of the
Lagrangian over $Y$ is finite exactly when
$\norm{\phi_i-\phi_j}_2\leq1$ for every pair. In that case, the infimum equals
$\inner{G}{\Phi}_F$. The Lagrange dual is therefore
\[
\sup_{\Phi}\inner{G}{\Phi}_F
\qquad\text{subject to}\qquad
\norm{\phi_i-\phi_j}_2\leq1
\quad\text{for every }i<j.
\]
The constraint is precisely $D(\Phi)\leq1$.

Equip the token set with the discrete metric, which assigns distance one to
every pair of distinct tokens. The constraint $D(\Phi)\leq1$ then says that
the map $i\mapsto\phi_i$ is vector valued and $1$-Lipschitz. The minimum-flow problem is a finite vector-valued Beckmann problem, while
the potential maximization is its vector-valued Kantorovich--Rubinstein dual
\citep{ciosmak2021optimaltransportvectormeasures,
robertson2025generalizationwassersteindistancebeckmann}.
Related formulations using Lipschitz-free spaces and tensor products appear
in \citet{arens1956embedding,CHAVEZDOMINGUEZ2011387,
cabrerapadilla2015lipschitztensorproduct}. Here, ``primal'' and ``dual'' refer to the flow problem and its Lagrange dual, respectively. Relative to the original diameter-constrained update problem, the minimum-flow problem provides its dual characterization.

\paragraph{Why the lower bound is exact.}
Write the flow problem as the second-order cone program
\[
\min_{Y,\{t_{ij}\}}
\sum_{i<j}t_{ij}
\quad\text{subject to}\quad
\sum_{j\neq i}y_{ij}=g_i,
\qquad
\norm{y_{ij}}_2\leq t_{ij}.
\]
Starting from the balanced flow constructed above, choose
$t_{ij}>\norm{y_{ij}^{(0)}}_2$ for every pair. This gives a Slater point
relative to the affine balance constraints. Hence the flow problem and its
dual have the same optimal value.

\paragraph{Common row translations.}
The potential $\Phi$ is defined only up to a common row translation. Let
$\bar\phi=V^{-1}\sum_i\phi_i$ and
$\widetilde\Phi=\Phi-\one\bar\phi^\top$. Then
\[
D(\widetilde\Phi)=D(\Phi),
\qquad
\inner{G}{\widetilde\Phi}_F
=
\inner{G}{\Phi}_F
-
\inner{\sum_i g_i}{\bar\phi}
=
\inner{G}{\Phi}_F.
\]
We may therefore impose $\one^\top\Phi=0$ without changing the dual value.
The centered feasible set is closed and bounded, hence compact. To see
boundedness, if $\sum_j\phi_j=0$ and $D(\Phi)\leq1$, then
\[
\begin{aligned}
\norm{\phi_i}_2
&=
\norm{\frac{1}{V}\sum_j(\phi_i-\phi_j)}_2 \\
&\leq
\frac{1}{V}\sum_j\norm{\phi_i-\phi_j}_2
\leq
\frac{V-1}{V}
\leq1.
\end{aligned}
\]
The dual maximum is therefore attained, and strong duality gives
\[
\mathcal T(G)
=
\max_{\substack{D(\Phi)\leq1\\\one^\top\Phi=0}}
\inner{G}{\Phi}_F.
\]

\paragraph{The diameter oracle.}
Suppose first that $\eta>0$. The substitution $S=-\eta\Phi$ is a
bijection between the centered potentials satisfying $D(\Phi)\leq1$ and
the centered updates satisfying $D(S)\leq\eta$. Therefore,
\[
\min_{\substack{D(S)\leq\eta\\\one^\top S=0}}
\inner{G}{S}_F
=
-\eta
\max_{\substack{D(\Phi)\leq1\\\one^\top\Phi=0}}
\inner{G}{\Phi}_F
=
-\eta\mathcal T(G).
\]
When $\eta=0$, the diameter and centering constraints force $S=0$, so both
sides are zero. This proves equation~\ref{eq:diameter-flow-duality} for
every $\eta\geq0$.

\paragraph{The active-edge condition.}
Let $Y^\star$ be a flow optimizer and let $\Phi^\star$ be a centered
potential optimizer. Set $S^\star=-\eta\Phi^\star$, which is an optimizer of
the centered diameter oracle. Strong duality, the balance equations, and the pairing identity in equation~\ref{eq:flow-pairing} give
\[
\begin{aligned}
0
&=
\mathcal T(G)-\inner{G}{\Phi^\star}_F \\
&=
\sum_{i<j}
\left(
\norm{y_{ij}^\star}_2
-
\inner{\phi_i^\star-\phi_j^\star}{y_{ij}^\star}
\right).
\end{aligned}
\]
Every summand is nonnegative because dual feasibility and
Cauchy--Schwarz give
\[
\norm{y_{ij}^\star}_2
-
\inner{\phi_i^\star-\phi_j^\star}{y_{ij}^\star}
\geq
\left(
1-\norm{\phi_i^\star-\phi_j^\star}_2
\right)
\norm{y_{ij}^\star}_2
\geq0.
\]
Since their sum is zero, every summand must vanish. Thus, whenever
$y_{ij}^\star\neq0$, equality throughout requires
\[
\phi_i^\star-\phi_j^\star
=
\frac{y_{ij}^\star}{\norm{y_{ij}^\star}_2}.
\]
Using $S^\star=-\eta\Phi^\star$ therefore gives
\begin{equation}
y_{ij}^\star\neq0
\quad\Longrightarrow\quad
s_i^\star-s_j^\star
=
-\eta\frac{y_{ij}^\star}{\norm{y_{ij}^\star}_2}.
\label{eq:cable-complementarity}
\end{equation}
Thus every edge used by an optimal flow is tight at the diameter bound.
Other token pairs may also be tight.

\paragraph{Mechanical interpretation and computational cost.}
The balance equation may be written as
\[
-g_i+\sum_{j\neq i}y_{ij}=0.
\]
Thus $-g_i$ can be viewed as an external force applied at token $i$, while the
vectors $y_{ij}$ are internal forces that balance it
\citep{ciosmak2021matrixholdersinequalitydivergence}. This is the source of
the term ``tension.'' Equation~\ref{eq:cable-complementarity} says that every
pair carrying nonzero tension is stretched to the full diameter in the
direction opposite to that tension.

The direct flow formulation contains one $d$-dimensional variable for each of
the $\binom{V}{2}$ token pairs. An optimal flow may be sparse, but the
formulation must allow every pair and therefore contains $O(V^2d)$ scalar
variables. This makes a direct solution unsuitable for every training step
and motivates the $O(Vd)$ RowNorm approximation in the main text.

\section{Proof of the projected RowNorm guarantee}
\label{app:rownorm-proof}

In this section, we prove equations~\ref{eq:projected-rownorm},
and the guarantee in equation~\ref{eq:rownorm-guarantee}, including sharpness of the latter.
Let $\one^\top G=0$, define
$\Sigma_G=\sum_i\norm{g_i}_2$, and recall that $PG=G$.

\paragraph{The radius-constrained problem.}
Fix $\eta\geq0$ and suppose $\operatorname{rad}(S)\leq\eta/2$, and let $\zeta$ be the center of
an enclosing ball of radius $\eta/2$. Since $\sum_i g_i=0$,
\[
\inner{G}{S}_F
=
\sum_i\inner{g_i}{s_i-\zeta}
\geq
-\sum_i\norm{g_i}_2\norm{s_i-\zeta}_2
\geq
-\frac{\eta}{2}\Sigma_G.
\]
Define
\[
\widehat S=-\frac{\eta}{2}\mathsf{RN}(G).
\]
Its rows lie in the ball of radius $\eta/2$ centered at the origin, and the
preceding inequalities hold with equality:
\[
\inner{G}{\widehat S}_F
=
-\frac{\eta}{2}\Sigma_G.
\]

The matrix $\widehat S$ need not have zero row mean, so we replace it by
$P\widehat S$. Writing $\bar s=V^{-1}\sum_i\widehat s_i$, the rows of
$P\widehat S$ are $\widehat s_i-\bar s$. A common translation changes
neither the covering radius nor the objective:
\[
\operatorname{rad}(P\widehat S)
=
\operatorname{rad}(\widehat S),
\qquad
\inner{G}{P\widehat S}_F
=
\inner{PG}{\widehat S}_F
=
\inner{G}{\widehat S}_F.
\]
Moreover, $\one^\top P\widehat S=0$, and, since $PG=G$,
\[
P\widehat S
=
-\frac{\eta}{2}P\,\mathsf{RN}(PG)
=
S_{\mathrm{RN}}(G;\eta).
\]
Therefore,
\[
S_{\mathrm{RN}}(G;\eta)
\in
\arg\min_{\substack{\operatorname{rad}(S)\leq\eta/2\\
                    \one^\top S=0}}
\inner{G}{S}_F,
\qquad
\Delta_{\mathrm{RN}}
=
\frac{\eta}{2}\Sigma_G.
\]
Since $D(S)\leq2\operatorname{rad}(S)$, projected RowNorm is also feasible
for the original diameter-constrained problem.

\paragraph{Reduction to the gradient span.}
For the comparison, assume $G\neq0$ and $\eta>0$. Let
\[
E=\operatorname{span}\{g_1,\ldots,g_V\},
\qquad
k=\dim(E)=\rank(G).
\]
For any feasible $S$, let $S_E$ have rows $(\Pi_Es_i)^\top$, where
$\Pi_E$ is the orthogonal projection onto $E$. Because each $g_i$ belongs
to $E$, projection preserves the objective, contracts every pairwise
distance, and preserves centering:
\[
\inner{G}{S_E}_F=\inner{G}{S}_F,
\qquad
\norm{\Pi_Es_i-\Pi_Es_j}_2
\leq
\norm{s_i-s_j}_2,
\qquad
\sum_i\Pi_Es_i
=
\Pi_E\!\left(\sum_i s_i\right)
=
0.
\]
The exact diameter problem may therefore be restricted to row clouds in the
$k$-dimensional space $E$.

\paragraph{Comparison with the diameter oracle.}
Jung's theorem states that every subset of a $k$-dimensional Euclidean space
with diameter at most $\eta$ lies in a ball of radius at most
$J_k\eta$, where
\[
J_k=\sqrt{\frac{k}{2(k+1)}}
\]
\citep{Jung1901,article}. Hence, for every feasible
update in $E$, there is a center $\zeta$ satisfying
$\norm{s_i-\zeta}_2\leq J_k\eta$ for every $i$. Using
$\sum_i g_i=0$ gives
\[
-\inner{G}{S}_F
=
-\sum_i\inner{g_i}{s_i-\zeta}
\leq
\sum_i\norm{g_i}_2\norm{s_i-\zeta}_2
\leq
J_k\eta\Sigma_G.
\]
In particular,
\[
\Delta_\star\leq J_k\eta\Sigma_G,
\qquad
\Delta_{\mathrm{RN}}=\frac{\eta}{2}\Sigma_G.
\]
Projected RowNorm is feasible for the diameter constraint, so
$\Delta_{\mathrm{RN}}\leq\Delta_\star$. Since
$G\neq0$ and $\eta>0$, both decreases are positive, and therefore
\[
\boxed{
\sqrt{\frac{k+1}{2k}}
=
\frac{1}{2J_k}
\leq
\frac{\Delta_{\mathrm{RN}}}{\Delta_\star}
\leq
1.
}
\]

\paragraph{Sharpness.}
Let $v_1,\ldots,v_{k+1}$ be the vertices of a centered regular
$k$-simplex with unit edge length. Then
\[
\sum_{i=1}^{k+1}v_i=0,
\qquad
\norm{v_i}_2=J_k.
\]
For $V=k+1$, take $g_i=v_i$ and $s_i^\star=-\eta v_i$. This update is
centered, has diameter $\eta$, and achieves
\[
-\inner{G}{S^\star}_F
=
\eta\sum_i\norm{v_i}_2^2
=
\eta(k+1)J_k^2
=
\eta J_k\Sigma_G.
\]
It attains the Jung upper bound and is therefore optimal. Since the normalized
simplex rows also have zero mean, projected RowNorm achieves
\[
\Delta_{\mathrm{RN}}
=
\frac{\eta}{2}\Sigma_G.
\]
Consequently,
\[
\frac{\Delta_{\mathrm{RN}}}{\Delta_\star}
=
\frac{1}{2J_k}
=
\sqrt{\frac{k+1}{2k}}.
\]

For $V>k+1$, append $V-k-1$ zero-gradient rows and place their update rows at
the simplex center. Since $J_k\leq1$, these additional rows do not increase
the diameter. The gradient and update remain centered, and neither objective
value changes. Thus the lower bound is sharp for every $V\geq k+1$ under the
fixed $\eta/2$ scaling in equation~\ref{eq:projected-rownorm}.

\end{document}